\pdfoutput=1

\documentclass[11pt]{article}

\usepackage[final]{acl}

\usepackage{times}
\usepackage{latexsym}

\usepackage[T1]{fontenc}

\usepackage[utf8]{inputenc}

\usepackage{microtype}

\usepackage{inconsolata}

\usepackage{graphicx}
\usepackage{xcolor}
\usepackage{tikz}

\definecolor{MiyaColorB}{RGB}{210,210,210} 
\definecolor{MiyaColorE}{RGB}{0,100,255} 
\tikzset{
	Miyacube/.style = {x={(1cm,0cm)},
		y={(0cm,1cm)},
		z={({0.4*cos(30)},{0.4*sin(30)})}, 
		scale=0.22},
	b/.style={draw, thick, darkgray, rectangle, rounded corners=2ex,minimum height=3em, minimum width=12em,align=left,text width=12em,inner sep=0.8ex,text=black,}, 
        bl1/.style={draw, thick, darkgray, rectangle, rounded corners=2ex,minimum height=3.5em, minimum width=10em,align=left,text width=10em,inner sep=0.8ex,text=black,}, 
        bl2/.style={draw, thick, darkgray, rectangle, rounded corners=2ex,minimum height=3em, minimum width=6em,align=left,text width=6em,inner sep=0.8ex,text=black,}, 
        threel/.style={draw, thick, darkgray, rectangle, rounded corners=2ex,minimum height=3.5em, minimum width=7.6em,align=left,text width=7.6em,inner sep=0.8ex,text=black,}, 
        sm/.style={draw, thick, darkgray, rectangle, rounded corners=2ex,minimum height=3em, minimum width=6em,align=left,text width=6em,inner sep=0.8ex,text=black,}, 
        tn/.style={draw, thick, darkgray, rectangle, rounded corners=2ex,minimum height=3em, minimum width=4em,align=left,text width=4em,inner sep=0.8ex,text=black,}, 
        tn2/.style={draw, thick, darkgray, rectangle, rounded corners=2ex,minimum height=3em, minimum width=5em,align=left,text width=5em,inner sep=0.8ex,text=black,}, 
        big1/.style={draw, thick, darkgray, rectangle, rounded corners=2ex,minimum height=28em, minimum width=34em,align=left,text width=34em,inner sep=0.8ex,text=black,}, 
        big/.style={draw, thick, darkgray, rectangle, rounded corners=2ex,minimum height=25em, minimum width=28em,align=left,text width=28em,inner sep=0.8ex,text=black,}, 
        big2/.style={draw, thick, darkgray, rectangle, rounded corners=2ex,minimum height=26em, minimum width=19em,align=left,text width=19em,inner sep=0.8ex,text=black,}, 
        med/.style={draw, thick, darkgray, rectangle, rounded corners=2ex,minimum height=15em, minimum width=15em,align=left,text width=15em,inner sep=0.8ex,text=black,}, 
        thin/.style={draw, thick, darkgray, rectangle, rounded corners=2ex,minimum height=5em, minimum width=4em,align=left,text width=4em,inner sep=0.8ex,text=black,}, 
	Miyaarrow/.style args={#1 colored by #2}{ 
		-{Triangle[scale=0.5]}, %
		line width=#1,#2,shorten <=(#1),
	}
} 
\title{Auto-SLURP: A Benchmark Dataset for Evaluating Multi-Agent Frameworks in Smart Personal Assistant
}

\author{Lei Shen \\
  GEB Tech\\
  \texttt{lorashen17@gmail.com} \\\And
  Xiaoyu Shen \\
  Ningbo Institute of Digital Twin, EIT, Ningbo\\
  \texttt{xyshen@eitech.edu.cn} \\}

\begin{document}
\maketitle
\begin{abstract}
In recent years, multi-agent frameworks powered by large language models (LLMs) have advanced rapidly. Despite this progress, there is still a notable absence of benchmark datasets specifically tailored to evaluate their performance. To bridge this gap, we introduce \textbf{Auto-SLURP}, a benchmark dataset aimed at evaluating LLM-based multi-agent frameworks in the context of intelligent personal assistants. Auto-SLURP extends the original SLURP dataset---initially developed for natural language understanding tasks---by relabeling the data and integrating simulated servers and external services. This enhancement enables a comprehensive end-to-end evaluation pipeline, covering language understanding, task execution, and response generation. Our experiments demonstrate that Auto-SLURP presents a significant challenge for current state-of-the-art frameworks, highlighting that truly reliable and intelligent multi-agent personal assistants remain a work in progress. The dataset and related code are available at \url{https://github.com/lorashen/Auto-SLURP/}.
\end{abstract}

\section{Introduction}

Multi-agent frameworks built on large language models (LLMs) have seen rapid development in recent years~\cite{li2023camel,su2024unraveling, hong2024metagpt, Wu2023AutoGenEN, liu2024agentlite}. These frameworks provide general-purpose infrastructures that facilitate the construction of multi-agent systems through modular architectures, communication protocols, and coordination strategies. Despite this progress, there remains a noticeable gap in standardized benchmarks tailored to evaluate the effectiveness of these frameworks.

While a number of benchmarks have been proposed to assess the tool-use capabilities of LLMs~\cite{qin2023toolllm, chen2023t, zhu2023weaker,zhuang2024toolqa, ye2024tooleyes}, they primarily focus on individual LLMs and address only a narrow slice of functionality. As a result, they do not adequately reflect the complexity, interactivity, and coordination challenges inherent in real-world multi-agent scenarios.

To capture broader dimensions of agent behavior, several social and interactive benchmarks have recently been proposed. For example, Cooperation~\cite{Abdelnabi2023CooperationCA}, SOTOPIA~\cite{zhou2024sotopia}, AgentSense~\cite{mou2024agentsense}, and SocialBench~\cite{chen2024roleinteract} create social environments to evaluate agents' interpersonal and collaborative abilities. In parallel, AgentBench~\cite{liu2023agentbench} targets reasoning and decision-making skills in domains such as coding, web navigation, and e-commerce. Other works, including MAgIC~\cite{xu2023magic}, CUISINEWORLD~\cite{gong-etal-2024-mindagent}, BattleAgentBench~\cite{wang2024battleagentbench}, CivRealm~\cite{qi2024civrealm}, and LegalAgentBench~\cite{li2024legalagentbench}, introduce game-based or domain-specific settings to assess multi-agent interaction and domain expertise.

Meanwhile, benchmarks in embodied environments—such as AgentBoard~\cite{ma2024agentboard}, ALFWorld~\cite{shridharalfworld}, the ThreeDWorld Transport Challenge~\cite{gan2021threedworld}, and WAH~\cite{puig2020watch}—focus on grounding agents in physical or simulated worlds. However, these efforts are typically designed to evaluate individual agents’ task execution and interaction capabilities, rather than to assess the performance or flexibility of open-source multi-agent frameworks. Moreover, the highly integrated nature of game-based and embodied environments often makes them difficult to adapt for evaluating general-purpose frameworks, limiting their reusability and extensibility~\cite{xu2020data,zhang2021knowledge}.


\begin{table*}
  \centering
  \begin{tabular}{lll}
    \hline
    User     & \multicolumn{2}{l}{could you please email john saying i'm on leave}
                 \\
    \hline
    \multicolumn{1}{l|}{}&\multicolumn{1}{l|}{\textbf{re-labeled}}&\textbf{original}\\
    \hline
    
    \multicolumn{1}{l|}{Intent}     & \multicolumn{1}{l|}{email\_sendemail}     &   email\_sendemail     \\
    \multicolumn{1}{l|}{Slots}     & \multicolumn{1}{l|}{to\_person: john, content: i'm on leave}&person : john\\
    \hline
  \end{tabular}
  \caption{The example of the annotations in Auto-SLURP.}
  \label{tab:sample}
\end{table*}

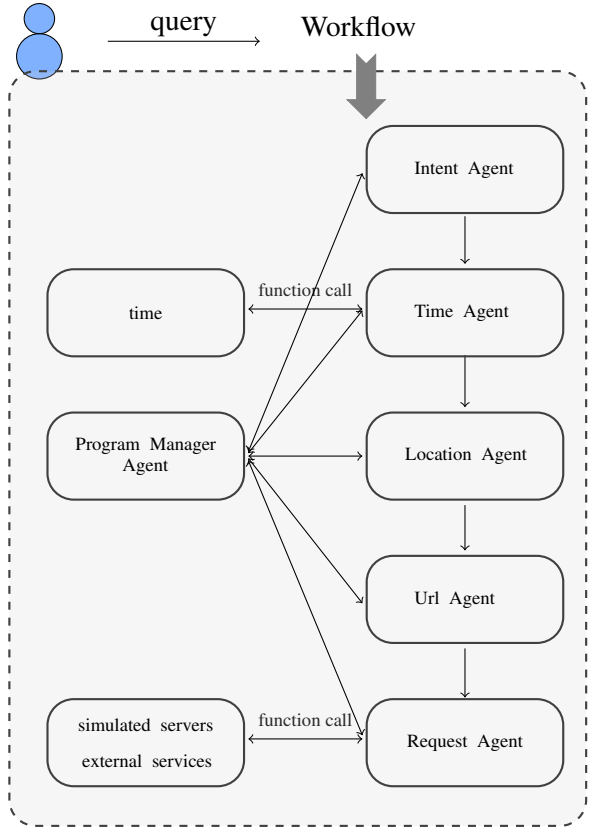
\begin{figure}[t]
		\centering
		\vskip 1ex
			\strut\vspace*{-\baselineskip}\\ 
                \scalebox{1.0}{
			\begin{tikzpicture}
   \node (e) at (-3.0,5.6) {};
   \draw[fill opacity=0.5,fill=MiyaColorE] (e) circle (0.2cm);
   \node (f) at (-3.0,5.1) {};
   \draw[fill opacity=0.5,fill=MiyaColorE] (f) circle (0.3cm);
   \node (block10) [xshift=-11mm,yshift=55mm]{query};
   \node (block10) [xshift=12mm,yshift=55mm]{Workflow};
   \draw[->](-2.1,5.3)-- (-0.1,5.3);
   \node (block1) [xshift=5mm,yshift=20mm]{\scriptsize function call};
   \node (block1) [xshift=5mm,yshift=-37mm]{\scriptsize function call};

   \node (block9) [big2,dashed,fill=MiyaColorB,fill opacity=0.2,xshift=4mm,yshift=-1mm] {
			};
  
			\node (block1) [sm,xshift=-16mm,yshift=-2mm] {\centering\scriptsize  Program Manager Agent

			};
		
                \draw[fill,gray] (1.3,5.0) --++ (.5 * .25,.5 * .25) -- node[above=.5cm,black,xshift = .25cm] {} ++ (0,-.6) --++ (.5 * .25,0) --++ (-1 * .25,-1 * .25) --++ (-1 * .25,1 * .25) --++ (.5 * .25,0) --++ (0,.6) --cycle;

            
            \node (block1_2) [bl2,xshift=26mm,yshift=36mm] {
            \scriptsize 
                \hspace{2.0em}Intent Agent

			};

            \draw[->](2.6,2.98)-- (2.6,2.35);
            \draw[->](2.6,1.14)-- (2.6,0.45);
            \draw[->](2.6,-0.85)-- (2.6,-1.45);
            \draw[->](2.6,-2.75)-- (2.6,-3.40);
            
            \draw[<->](-0.25,-0.14)-- (1.25,3.55);
            \draw[<->](-0.25,-0.16)-- (1.25,1.73);
            \draw[<->](-0.25,-0.20)-- (1.25,-0.2);
            \draw[<->](-0.25,-0.23)-- (1.25,-2.15);
            \draw[<->](-0.25,-0.26)-- (1.25,-3.9);
            
            \draw[<->](-0.25,1.75)--(1.25,1.75);
            \draw[<->](-0.25,-3.95)--(1.25,-3.95);
			
		\node (block15) [sm,xshift=-16mm,yshift=17mm] 
            {\centering\scriptsize
                \hspace{3.8em}time 
};
			
  \centering
				
  \centering
 
                \node (block12) [bl2,xshift=26mm,yshift=17mm] {\centering\scriptsize 
                \hspace{2.0em}Time Agent
                };

                \node (block12) [bl2,xshift=26mm,yshift=-2mm] {\centering\scriptsize 
                \hspace{1.5em}Location Agent
                };

                \node (block15) [sm,xshift=-16mm,yshift=-40mm] {\centering\scriptsize
                \hspace{0.0em}simulated servers 
                
                \hspace{1.3em}external services
};

                \node (block14) [bl2,xshift=26mm,yshift=-21mm] {\centering\scriptsize 
                \hspace{2.0em}Url Agent
            };
			\node (block3) [bl2,xshift=26mm,yshift=-40mm] {\centering\scriptsize 
                \hspace{-0.5em}
                Request Agent
               
};

			\end{tikzpicture}
		}
  \caption{\small The workflow defined for the Auto-SLURP dataset.}
    \label{fig:iot}
	\end{figure}

Taken together, although significant progress has been made in benchmarking agent capabilities, existing efforts do not sufficiently address the unique needs of evaluating multi-agent frameworks. This highlights a pressing need for a comprehensive and flexible benchmark that can rigorously and fairly assess the effectiveness of LLM-based multi-agent infrastructures across a range of scenarios.

The vision of an intelligent personal assistant—an AI system capable of understanding natural language and performing tasks on behalf of users—has long captured the imagination of both researchers and the public~\cite{edu2020smart,Hoy02012018}. Despite significant progress in AI and the emergence of powerful LLM-based multi-agent systems, this vision remains underexplored in the context of multi-agent evaluation. Personal assistants are expected to handle a wide range of tasks, such as checking the weather, sending emails, managing calendars, and controlling IoT devices. Achieving this level of functionality demands not only natural language understanding (NLU), but also sophisticated capabilities in decision-making, reasoning, tool use, coordination, and adaptability~\cite{del2021question,shen2022low}.

To help fill this gap, we introduce \textbf{Auto-SLURP}, a benchmark designed to evaluate the effectiveness of LLM-based multi-agent frameworks in building intelligent personal assistants. Auto-SLURP is built upon the SLURP dataset~\cite{bastianelli2020slurp,liu2021benchmarking}, originally created for natural language understanding in smart home scenarios. We extend SLURP’s original intent-slot structure to support comprehensive end-to-end evaluation: from language understanding and intent interpretation, to task execution and response generation. To better reflect the complexity of real-world interactions, we relabel the slots and restructure the data to align with complete user-interaction pipelines.

Auto-SLURP simulates realistic assistant interactions by integrating external services and simulated servers, enabling thorough evaluation of a framework’s ability to handle complex, multi-step operations. These operations include API access, state management across modules, and coordination between agents with specialized responsibilities. This setup allows us to assess not just whether multi-agent frameworks can interpret user commands, but also whether they can effectively orchestrate the backend processes needed to carry them out.

The dataset spans a wide range of task domains, such as calendar management, media playback, transportation scheduling, and information retrieval. This diversity ensures that Auto-SLURP serves as a robust and representative benchmark for evaluating both the flexibility and reliability of multi-agent frameworks in realistic scenarios.
Our experimental results demonstrate that Auto-SLURP presents significant challenges even for state-of-the-art multi-agent frameworks. These findings underscore the complexity involved in achieving seamless, intelligent assistant behavior and reveal that we are still some distance away from building fully dependable AI-based personal assistants.

\begin{table*}
  \centering
  \begin{tabular}{lcccc}
    \hline
     & \textbf{CamelAI}  & \textbf{LangGraph}  & \textbf{AutoGen} & \textbf{AgentLite}\\
    \hline
    acc     & 0.21    &    0.32    &    0.44    &    0.46           \\
    \hline
  \end{tabular}
  \caption{\small The results of the multi-agent frameworks.}
  \label{tab:acc}
\end{table*}

\begin{table*}
  \centering
  \begin{tabular}{lcccc}
    \hline
     & \textbf{CamelAI}  & \textbf{LangGraph}  & \textbf{AutoGen} & \textbf{AgentLite}\\
    \hline
    intent     & 54\%    &    34\%    &    68\%    &    69\%           \\
    time     & 18\%    &    12\%    &    9\%    &    19\%           \\
    location     & -    &    -    &    -    &    7\%           \\
    url     & 14\%    &    13\%    &    43\%    &    19\%           \\
    manager     & 9\%    &    53\%    &    13\%    &    -           \\
    function\_call&18\%    &    -    &    -    &    -            \\
    \hline
  \end{tabular}
  \caption{\small Failure reasons of the frameworks. Because one failure can be caused by multiple reasons, they do not sum up to 100\%.}
  \label{tab:reason}
\end{table*}
\section{Dataset Construction}

\paragraph{Creation of queries and annotations}

We make modification to the SLURP dataset, which is collected for the development of smart personal assistants. Personal assistant systems are inherently complex, as they must interpret and respond to a wide variety of user commands. SLURP was initially released for natural language understanding tasks~\cite{weld2022survey,yang2017end,shen2017estimation,su2018dialogue,huang2021dependency}, with a focus on intention detection and slot filling. In traditional methods, intent detection is treated as a classification problem, while slot filling is handled as a sequence-to-sequence task. For example, given the user query "play kari jobe for me", the intent is identified as "play\_music", and the slot is "artist\_name: kari jobe". In the original dataset, the slots are limited to the entities explicitly mentioned in the utterance, omitting other crucial information required to successfully execute the command. This omission can lead to incomplete or failed task execution.

To adapt SLURP for our specific use case, we retain only the user queries and their corresponding intents from SLURP, while re-labeling the slots. Specifically, we enrich the slot information by adding new slots and refining existing ones to capture all the information necessary for backend task execution. We also ensure that the slot structures are compatible with LLMs, which typically generate outputs rather than classify them. Table \ref{tab:sample} illustrates an example of our modified samples, with our re-labeled version in the middle column, and the original SLURP sample in the right column.

The dataset encompasses a wide range of tasks, from straightforward actions like setting calendars or playing music, to more complex operations such as information retrieval or handling transportation-related commands. We randomly select 1,000 samples from the training set and 100 samples from the testing set. Based on our experimental results, this subset is considered sufficient for training and testing LLM-based multi-agent frameworks.

\paragraph{Collection of the end servers}

To evaluate end-to-end system performance, we simulate the execution servers that process and carry out user commands. This simulation enable us to verify whether the commands are correctly interpreted and executed, ensuring that the overall system functions as expected. In our training set, we identify 23 distinct domains. For each domain, we build a dedicated server to handle the relevant operations.
Additionally, for certain domains which require external information, such as search, weather, and news, we integrate external services, i.e., third-party APIs. These API calls allow the system to fetch the required information, ensuring that user requests are handled efficiently and with up-to-date content.
\begin{table*}
  \centering
  \begin{tabular}{lcccc}
    \hline
     USD/query & \textbf{CamelAI}  & \textbf{LangGraph}  & \textbf{AutoGen} & \textbf{AgentLite}\\
    \hline
    cost     & 0.52    &    0.14    &    0.80    &    0.55           \\
    \hline
  \end{tabular}
  \caption{The costs of the frameworks.}
  \label{tab:cost}
\end{table*}

\section{Experiments}

\subsection{Setup}

We compare several representative LLM-based multi-agent frameworks.

\noindent\textbf{CamelAI} \cite{li2023camel} introduces a cooperative multi-agent framework that allows communicative agents to autonomously collaborate toward completing tasks through role-playing.

\noindent\textbf{AutoGen} \cite{Wu2023AutoGenEN} presents a customizable multi-agent conversation framework that can integrate LLMs, humans, and tools, enabling dynamic agent interactions.

\noindent\citet{graph} is built upon the foundation of \citet{chain} and provides an easy way to create cyclical graphs, which is particularly useful for creating agent runtimes.

\noindent\textbf{AgentLite}~\cite{liu2024agentlite} is a lightweight, modular codebase for developing customized LLM-based agent systems. It enables researchers to easily build prototype applications and experiment with new reasoning strategies and agent architectures.

For all multi-agent frameworks, we use GPT-4~\cite{Achiam2023GPT4TR} as the LLM. The prompts are created and adjusted during the setup phrase. The temperature is set as 0 to ensure that the LLM's responses are deterministic and fixed.

\subsection{Defined workflows}
We use each multi-agent framework to build a workflow that simulates a smart personal assistant. In the workflow, a program manager agent serves as the orchestrator; it processes the user's input query and delegates subtasks to specialized agents. We introduce an intent agent to predict the intent and slots. Additionally, we add a time agent and a location agent to format the time and location parameters, if applicable. We adopt a url agent to select the appropriate url from a list of candidates, and a request agent to execute the tool function call for the request. The overall workflow is illustrated in Figure \ref{fig:iot}. Although the orchestration methods, prompt policies, and reasoning approaches vary across frameworks, we ensure a fair and controlled comparison by maintaining consistency in the assigned roles, accessible tools, and prompts used to define agent functions during construction.
\subsection{Evaluation}
We use the successful execution rate as the evaluation metric, which measures the percentage of queries that are completed successfully from end to end. This metric assesses the reliability, efficiency, and ability of the framework to perform the intended actions without failure.
Additionally, we provide an automated evaluation tool that measures performance across all frameworks consistently and efficiently.

\begin{table}
  \centering
  \begin{tabular}{lcc}
    \hline
    \textbf{AutoGen} & \textbf{original}  & \textbf{finetuned}  \\
    \hline
    acc     & 0.40    &    0.62              \\
    \hline
  \end{tabular}
  \caption{The results for AutoGen with and without funetuning.}
  \label{tab:ft}
\end{table}

\section{Experiment Results}
\subsection{Results analysis}
Table~\ref{tab:acc} presents the results of the multi-agent frameworks.Among them, CamelAI achieves the lowest accuracy score, while AgentLite performs the best.
CamelAI's failure can be attributed to its difficulty in selecting the right tool to execute. LangGraph also underperforms, mainly because it only combines the system prompt and all the agents’ results into one list as input, without any adjustments. In contrast, AutoGen separates the prompts for the manager agent and the subtask agents, enabling clearer task delegation and yielding better results. AgentLite further improves performance by adopting "think and react" methods in the process, which significantly enhances execution success. Example prompts for LangGraph and AutoGen are provided in Appendix \ref{sec:appendixc}.
We also test other frameworks, such as AgentVerse~\cite{Chen2023AgentVerseFM} and AutoAgents~\cite{Chen2023AutoAgentsAF}. However, these frameworks either lack a generalized orchestration policy to support this scenario or do not provide sufficient information for effective implementation. This highlights the inherent complexity of designing robust multi-agent frameworks.

To gain deeper insight into failure points, we analyze the errors caused by individual agents and the function call part. As shown in Table \ref{tab:reason}, it is clear that the main source of failure stems from the intent agent. We show the failure attribution criteria in Appendix \ref{sec:appendixb}.

Furthermore, we analyze the cost of each framework. As shown in Table~\ref{tab:cost}, the costs are at the same level for CamelAI, AutoGen, and AgentLite, but LangGraph has a significantly lower cost. We believe this is because LangGraph only uses the system prompt and all agents’ results as input. Therefore, the cost for each query, ranging from 0.5 to 0.8, is reasonable for an advanced multi-agent framework in this scenario.

\subsection{Ablation}
Our earlier analysis demonstrates that most of the failures are caused by intent agent. To address this, we conduct an ablation study by further finetuning a model for the intent agent to assess its impact on overall framework performance.
We choose the open-source Llama 3 model~\cite{llama3modelcard} for finetuning. Specifically, we finetune the LLAMA-3 8B model using the training set and use the finetuned version as the intent agent. We report the results on AutoGen framework, and the results are listed in Table~\ref{tab:ft}. Compared to the framework that uses the original LLAMA-3 8B model, the finetuned version shows a performance improvement of 55\%. This result demonstrates that improving individual components—especially the main failure source—can significantly enhance the overall performance of multi-agent frameworks. A more detailed breakdown of domain-specific accuracy for both versions is provided in Appendix \ref{sec:appendix}.

Based on the analysis above, it is clear that we are still a few steps away from achieving a fully reliable and smart personal assistant. Achieving this goal will require continued progress in several key areas of multi-agent framework design—namely, the development of generalized orchestration policies, effective prompting methods, robust reasoning approaches (such as think and react), and careful selection of LLMs suited to the task.

\section{Conclusion}
We present Auto-SLURP, a dataset designed to evaluate LLM-based multi-agent frameworks. We assess the end-to-end execution tasks, not just the nature language understanding tasks. By incorporating simulated servers and external services, we evaluate the capacity of the frameworks to complete the entire process. The dataset proves to be sufficiently challenging to test the state-of-the-art multi-agent frameworks.

\section{Limitations}

The dataset incorporates simulated servers and external services, which may not fully mimic the behavior of real-world systems. This could result in discrepancies between the performance of frameworks in the benchmark and their performance in live applications.

Additionally, the dataset's evaluation is heavily reliant on the performance of LLMs. Variations in the quality and capabilities of LLMs across different versions could influence the outcomes.

\appendix
\section{Prompt Examples from LangGraph and AutoGen}
\label{sec:appendixc}

Below is an example prompt from LangGraph, which includes the agents’ names, the function description of the orchestration agent, the current subtask, and the responses from previous agents.

\noindent\{'content': 'You are a supervisor tasked with managing a conversation between the following workers to finish the first user's cmd: ['intent', 'time', 'location', 'url', 'request', 'genresponse']. Given the following user request, respond with the worker to act next. you are controlling smart home system, you have intent, time, location, and url agent and request to complete the user's task. You should first use intent to complete the intent prediction. Then if the result has time or location params, please try to ask time or location to solve the time and location. At last you should choose the url using url agent, and then use request to send and receive request to the url such as weather server and then use genresponse to generate response, then finalize the task. Even if the request's response is need further information or is a question, do not further answer the question, just finish the task. The response need to be the worker to act next, for example: \{"next": "FINISH"\}. When finished, respond with FINISH. the data in json.', 'role': 'system'\}, \{'content': 'will i need sunscreen this afternoon', 'role': 'user'\}, \{'content': 'domain:weather, intent:weather\_query, slots:time:this afternoon', 'name': 'intent', 'role': 'user'\}

The following is an example prompt from AutoGen, which includes a description of the overall task, detailed function descriptions of all agents, responses from previous agents, and the current subtask. (Some content has been omitted for brevity.)

\noindent\{'content': "You are in a role play game. The following roles are available: user\_proxy: A computer terminal that performs no other action than running Python scripts (provided to it quoted in python code blocks), or sh shell scripts (provided to it quoted in sh code blocks). Product\_manager: you are controlling smart home system, you have intent assistant, time\_assistant, location\_assistant, url\_assistant and request\_assistant to complete the user's task. You should first use intent to complete the intent prediction. Then if the result has time or location params, please try to ask time\_assistant or location\_assistant to solve the time and location. Then you choose the url using url\_assistant. At last you should use request\_assistant to send and receive request through functions from other servers such as weather server and response to user. You should generates reponse for the user, and tell manager to finalize the task. intent: Read the examples and results, and predict intent for the sentence. For 'set the alarm to two pm', first predict the domain, as domain:alarm, then the intent and slots, as the format: intent:alarm\_set,time:two pm. the intents are calendar: calendar\_set, calendar\_remove, calendar\_query ... Time\_assistant: Read the time params, and convert to formated time. If has date, call the user\_proxy\_auto get\_time function to get today's date, then calculate and format the date mentioned in the params. The time is 10:00. If has time, the time format should be 10:00. If no time specify, can return default time. If no date and time params, just skip. Location: Read the location params, and convert to formated location. The current location is new york. url\_assistant: Read the params, and choose the url from the servers' url list: qa server is ... then all the url format should be ... Request: for url and query params, use the request functions you have been provided with. Read the following conversation. Then select the next role from ['user\_proxy', 'Product\_manager', 'intent', 'Time\_assistant', 'Location', 'url\_assistant', 'Request'] to play. Only return the role.", 'role': 'system'\}, \{'content': '\{"query": "will i need sunscreen this afternoon"\}', 'role': 'user', 'name': 'user\_proxy'\}, \{'content': 'domain:weather,intent:weather\_query,time:this afternoon', 'role': 'user', 'name': 'intent'\}, \{'content': "Read the above conversation. Then select the next role from ['user\_proxy', 'Product\_manager', 'intent', 'Time\_assistant', 'Location', 'url\_assistant', 'Request'] to play. Only return the role.", 'name': 'checking\_agent', 'role': 'system'\}

\section{Failure Attribution Criteria in Evaluation}
\label{sec:appendixb}

During evaluation, the workflow proceeds even if a failure occurs, and task completion is assessed only after the entire process is complete. We then identify the agent responsible for the failure based on the following criteria:

\begin{itemize} 
\item Intent Agent: If the intent agent makes an incorrect prediction that ultimately leads to a workflow failure, the error is attributed to the intent agent. 
\item Time Agent: If the time agent provides an incorrect time that affects the final outcome, the error is assigned to the time agent. 
\item Location Agent: If the location agent supplies an incorrect location resulting in an incorrect outcome, the error is attributed to the location agent. 
\item URL Agent: If the URL agent selects the wrong URL or incorrect parameters, the error is considered to originate from the URL agent. Additionally, if the URL agent receives an incorrect intent but is capable of correcting it and fails to do so, the error is also attributed to the URL agent. 
\item Manager Agent: If the manager agent incorrectly selects the next agent in the workflow, causing a failure, the error is attributed to the manager agent. 
\item Function Call: If the system executes an incorrect function call that results in a failure, the error is classified as a function call failure. \end{itemize}

\section{Evaluation of Intent Prediction Accuracy Across Domains}
\label{sec:appendix}
\begin{table}[t]
  \centering
  \begin{tabular}{lcc}
    \hline
Domain	& Original &	Finetuned\\
\hline
Audiobook	& 0.0\%	& 66.7\%\\
Calendar	& 11.8\%	& 76.5\%\\
Currency	& 0.0\%	& 66.7\%\\
Datetime	& 14.3\%	& 71.4\%\\
Email	& 0.0\%	& 71.4\%\\
IoT	& 33.3\%	& 75.0\%\\
Lists	& 40.0\%	& 100.0\%\\
Music	& 0.0\%	& 70.0\%\\
News	& 0.0\%	& 100.0\%\\
Podcasts	& 0.0\%	& 50.0\%\\
QA	& 0.0\%	& 80.0\%\\
Radio	& 0.0\%	& 66.7\%\\
Recommendation	& 0.0\%	& 60.0\%\\
Transport	& 33.3\%	& 66.7\%\\
Weather	& 0.0\%	& 100.0\%\\
    \hline
  \end{tabular}
  \caption{The accuracy for each domain of the original model and the finetuned model.}
  \label{tab:domain}
\end{table}
We analyze the intent prediction accuracy across different domains, excluding those with fewer than three samples. The results are reported in Table~\ref{tab:domain}, showing the accuracy for each domain using both the original and the finetuned models.

To further investigate the performance gap between overall workflow accuracy and intent accuracy, we examine the outputs of the intent agent. We find that some errors from the original model were due to formatting issues—such as incorrect slot names or responses provided as plain-text descriptions rather than structured outputs. These cases can often be corrected by downstream agents (e.g., the URL agent) within the workflow. Therefore, for reference, we additionally report the intent accuracy when ignoring slot name errors in Table~\ref{tab:domain2}.

As shown in the two tables, finetuning leads to improved accuracy across all domains. Among them, the Podcasts domain appears to be more challenging for the intent agent, achieving a final accuracy of only 50.0\%.
\newpage
\begin{table}[ht]
  \centering
  \begin{tabular}{lcc}
    \hline
Domain	& Original &	Finetuned\\
\hline
Audiobook	&0.0\%	&66.7\%\\
Calendar	&29.4\%	&82.4\%\\
Currency	&0.0\%	&66.7\%\\
Datetime	&42.9\%	&85.7\%\\
Email	&57.1\%	&71.4\%\\
IoT	&58.3\%	&75.0\%\\
Lists	&40.0\%	&100.0\%\\
Music	&10.0\%	&70.0\%\\
News	&33.3\%	&100.0\%\\
Podcasts	&0.0\%	&50.0\%\\
QA	&0.0\%	&80.0\%\\
Radio	&0.0\%	&66.7\%\\
Recommendation	&20.0\%	&60.0\%\\
Transport	&66.7\%	&66.7\%\\
Weather	&14.3\%	&100.0\%\\
    \hline
  \end{tabular}
  \caption{The accuracy (ignore slot name errors) for each domain of the original model and the finetuned model.}
  \label{tab:domain2}
\end{table}

\end{document}